%% file: main.tex
\documentclass[11pt,a4paper]{article}

\usepackage[hyperref]{naaclhlt2019}
\usepackage{times}
\usepackage{latexsym}
\usepackage{url}

\usepackage{color}
\usepackage{multirow}
\usepackage{multicol}
\usepackage{commath}
\usepackage{amsmath,bm}
\usepackage{tabularx}
\usepackage{algpseudocode}
\usepackage{dsfont}
\usepackage{booktabs}
\usepackage{pbox}
\usepackage{graphicx}
\usepackage{enumitem}
\usepackage{framed}
\usepackage{array}
\usepackage[linesnumbered,ruled,vlined]{algorithm2e}
\usepackage[linesnumbered]{algorithm2e}
\usepackage[detect-all]{siunitx}

\usepackage{amssymb}
\newcommand{\nonl}{\renewcommand{\nl}{\let\nl\oldnl}}%

\aclfinalcopy %
\usepackage{xspace,mfirstuc,tabulary}

\newcommand{\eg}{{e.g.}}
\newcommand{\ie}{{i.e.}}
\newcommand{\vs}{{vs.}}
\newcommand{\hhide}[1]{}

\title{Improving Machine Reading Comprehension with \\General Reading Strategies}

\author{
  Kai Sun\textsuperscript{1}\Thanks{This work was done when K. S. was an intern at the Tencent AI Lab, Bellevue, WA.}
  ~~ Dian Yu\textsuperscript{2} ~~ Dong Yu\textsuperscript{2} ~~ Claire Cardie\textsuperscript{1}\\
 \textsuperscript{1}Cornell University, Ithaca, NY, USA \\
 \textsuperscript{2}Tencent AI Lab, Bellevue, WA, USA\\
   ks985@cornell.edu, \{yudian, dyu\}@tencent.com, cardie@cs.cornell.edu \\
}

\date{}

\begin{document}
\maketitle

\input{0_abstract.tex}           %
\input{1_introduction.tex}
%
\input{2_task.tex}               %
\input{3_method.tex}             %
\input{4_evaluation.tex}
%
\input{5_relatedwork.tex}        %
\input{6_conclusion.tex}
%

\section*{Acknowledgments}
We would like to thank the anonymous reviewers for their constructive suggestions. We thank Hai Wang, Chengzhu Yu, and Chao Weng for their useful discussions. We especially thank Chao for helping us speed up the release of the preprint\footnote{\url{https://arxiv.org/abs/1810.13441v1}.} with technical supports. We thank Rishi Bommasani for proofreading the paper and Saku Sugawara for sharing annotations with us.

\bibliography{naaclhlt2019}
\bibliographystyle{acl_natbib}

\end{document}

%% file: 0_abstract.tex
\begin{abstract}

Reading strategies have been shown to improve comprehension levels, especially
for readers lacking adequate prior knowledge. Just as the process of knowledge accumulation is time-consuming for human readers, it is resource-demanding to impart rich general domain knowledge into a deep language model via pre-training. Inspired by reading strategies identified in cognitive science, and given limited computational resources -- just a pre-trained model and a fixed number of training instances -- we propose three general strategies aimed to improve non-extractive machine reading comprehension (MRC): (i) {\sc Back and Forth Reading} that considers both the original and reverse order of an input sequence, (ii) {\sc Highlighting}, which adds a trainable embedding to the text embedding of tokens that are relevant to the question and candidate answers, and (iii) {\sc Self-Assessment} that generates practice questions and candidate answers directly from the text in an unsupervised manner. 

By fine-tuning a pre-trained language model~\cite{radfordimproving} with our proposed strategies on the largest general domain multiple-choice MRC dataset RACE, we obtain a $5.8\%$ absolute increase in accuracy over the previous best result achieved by the same pre-trained model fine-tuned on RACE without the use of strategies. We further fine-tune the resulting model on a target MRC task, leading to an absolute improvement of $6.2\%$ in average accuracy over previous state-of-the-art approaches on six representative non-extractive MRC datasets from different domains (\ie, ARC, OpenBookQA, MCTest, SemEval-2018 Task 11, ROCStories, and MultiRC). These results demonstrate the effectiveness of our proposed strategies and the versatility and general applicability of our fine-tuned models that incorporate these strategies. Core code is available at \url{https://github.com/nlpdata/strategy/}.

\end{abstract}

%% file: 1_introduction.tex
\section{Introduction}

Recent years have seen a growing interest in machine reading comprehension (MRC)~\cite{rajpurkar2016squad,choi2018quac,kovcisky2018narrativeqa,reddy2018coqa}. In this paper, we mainly focus on \textit{non-extractive MRC}~\cite{khashabi2018looking,ostermann2018semeval,clark2018think}, in which a significant percentage of candidate answers are not restricted to text spans from the reference document or corpus. In comparison to \textit{extractive MRC} tasks (Section~\ref{sec:task:extractive}), non-extractive MRC (Section~\ref{sec:task:multiple}) requires diverse reading skills and, as a result, the performance of machine readers on these tasks more accurately indicates the comprehension ability of machine readers in realistic settings such as exams~\cite{lai2017race}. 

Recently, significant progress has been achieved on many natural language processing tasks including MRC by fine-tuning a pre-trained general-purpose language model~\cite{radfordimproving,bert2018}. 
However, similar to the process of knowledge accumulation for human readers, it is time-consuming and resource-demanding to impart massive amounts of general domain knowledge from external corpora into a deep language model via pre-training. For example, it takes a month to pre-train a $12$-layer transformer on eight P$100$ GPUs over the BooksCorpus~\cite{zhu2015aligning,radfordimproving};~\newcite{bert2018} pre-train a $24$-layer transformer using $64$ TPUs for four days on the BooksCorpus plus English Wikipedia, a feat not easily reproducible considering the tremendous computational resources ($\approx$ one year to train on eight P$100$ GPUs).  %

From a practical viewpoint, given a limited number of training instances and a pre-trained model, can we improve machine reading comprehension during fine-tuning instead of imparting more prior knowledge into a model via expensive pre-training? Inspired by reading strategies identified in cognitive science research that have been shown effective in improving comprehension levels of human readers, especially those who lack adequate prior knowledge of the topic of the text~\cite{mokhtari2002measuring,mokhtari2002assessing,mcnamara2004istart}, we propose three corresponding domain-independent strategies to improve MRC based on an existing pre-trained transformer (Section~\ref{sec:trans}): %

\vspace{-\topsep}
\begin{itemize}
\setlength\itemsep{-0.25em}
    \item {\sc Back and Forth Reading} (\emph{``I go back and forth in the text to find relationships among ideas in it.''}):\\
    consider both the original and reverse order of an input sequence (Section~\ref{sec:fb})
    
    \item {\sc Highlighting} (\emph{``I highlight information in the text to help me remember it.''}):\\
    add a trainable embedding to the text embedding of those tokens deemed relevant to the question and candidate answers (Section~\ref{sec:hl})
    
    \item {\sc Self-Assessment} (\emph{``I ask myself questions I would like to have answered in the text, and then I check to see if my guesses about the text are right or wrong.''}):\\
    generate practice questions and their associated span-based candidate answers from the existing reference documents (Section~\ref{sec:sa})
\end{itemize}
\vspace{-\topsep}

By fine-tuning a pre-trained transformer~\cite{radfordimproving} according to our proposed strategies on the largest general domain multiple-choice MRC dataset RACE~\cite{lai2017race} collected from language exams, we obtain a $5.8\%$ absolute improvement in accuracy over the previous best result achieved by the same pre-trained transformer fine-tuned on RACE without the use of strategies (Section~\ref{sec:eval:race}). We further fine-tune the resulting model on a target MRC task. Experiments show that our method achieves new state-of-the-art results on six representative non-extractive MRC datasets that require a range of reading skills such as commonsense and multi-sentence reasoning (\ie, ARC~\cite{clark2016combining,clark2018think}, OpenBookQA~\cite{mihaylov2018can}, MCTest~\cite{richardson2013mctest}, SemEval-2018 Task 11~\cite{yang2017semi}, ROCStories~\cite{mostafazadeh2016corpus}, and MultiRC~\cite{khashabi2018looking}) (Section~\ref{sec:eval:five}). These results indicate the effectiveness of our proposed strategies and the versatility and generality of our fine-tuned models that incorporate the strategies.

%% file: 2_task.tex
\section{Task Introduction}
\label{sec:mrc}

We roughly categorize machine reading comprehension tasks into two groups: extractive (Section~\ref{sec:task:extractive}) and non-extractive (Section~\ref{sec:task:multiple}) based on the expected answer types.

\begin{table*}[htbp!]
\centering
\scriptsize
\begin{tabular}{lccccccc}
\toprule
&\textbf{RACE}   &\textbf{ARC}  &\textbf{OpenBookQA}    &\textbf{MCTest}  &\textbf{SemEval-2018 Task 11} &\textbf{ROCStories} &\textbf{MultiRC}\\
\midrule            
construction method    & exams   & exams      & crowd.     & crowd.    & crowd.    & crowd.  & crowd.            \\
sources of documents   & general & science    & science   &  stories  & narrative text   & stories & mixed-domain    \\
average \# of answer options & 4.0  & 4.0     & 4.0       & 4.0       & 2.0       & 2.0    & 5.4             \\
\# of documents     &  27,933    & 14M$^\dagger$ & 1,326$^\dagger$    & 660       &  2,119   & 3,742 & 871   \\
\# of questions     &  \bf 97,687     & 7,787      & 5,957     & 2,640   &   13,939       &  --  &   9,872           \\

\midrule
non-extractive answer$^\star$ (\%)       & 87.0      & 43.3   & 83.8   & 45.3   &   89.9  &  100.0 & 82.1      \\
\bottomrule
\end{tabular}
\caption{Statistics of multiple-choice machine reading comprehension datasets. Some values come from~\newcite{reddy2018coqa},~\newcite{kovcisky2018narrativeqa}, and~\newcite{lai2017race} (crowd.: crowdsourcing; $^\dagger$: regarding each sentence/claim as a document~\cite{clark2018think}; $^\star$: correct answer options that are not text snippets from reference documents).}
\label{tab:related:answer}
\end{table*}

\subsection{Extractive MRC}
\label{sec:task:extractive}
Recently large-scale extractive MRC datasets have been constructed~\cite{hermann2015teaching,hill2015goldilocks,onishi2016did,chen2016character,mostafazadeh2016corpus,bajgar2016embracing,nguyen2016ms,triviaQA,ma2018challenging}, such as SQuAD~\cite{rajpurkar2016squad} and NewsQA~\cite{trischler2017newsqa}. Given a reference document and a question, the expected answer is a short span from the document. In contrast, answers in datasets such as SearchQA~\cite{dunn2017searchqa} and NarrativeQA~\cite{kovcisky2018narrativeqa} are free-form human generated texts based on given documents~\cite{nguyen2016ms,reddy2018coqa,choi2018quac}. However, since annotators tend to directly copy spans as answers, the majority of answers are still extractive~\cite{reddy2018coqa,kovcisky2018narrativeqa}. %
\subsection{Non-Extractive MRC}
\label{sec:task:multiple}
In this section, we primarily discuss multiple-choice MRC datasets, in which answer options are not restricted to extractive text spans. Given a question and a reference document/corpus, multiple answer options are provided, and at least one of them is correct. It involves extensive
human efforts to build such a dataset (\eg, MCTest~\cite{richardson2013mctest}, SemEval-$2018$ Task $11$~\cite{ostermann2018semeval}, MultiRC~\cite{khashabi2018looking}, and OpenBookQA~\cite{mihaylov2018can}) by crowdsourcing. %
Besides crowdsourcing, datasets such as RACE~\cite{lai2017race} and ARC~\cite{clark2018think} are collected from language or science exams designed by educational experts~\cite{penas2014overview,shibuki2014overview,tseng2016towards} to evaluate the comprehension level of human participants. Compared to questions in extractive MRC tasks, besides surface matching, there are various types of complicated questions such as math word problems, summarization, logical reasoning, and sentiment analysis, requiring advanced reading skills and prior world knowledge. Besides, in most cases, we can adopt an objective evaluation criteria such as accuracy to evaluate system performance~\cite{clark2016combining,lai2017race}. As these kind of datasets are relatively difficult to construct or collect, most existing datasets are small in size, which hinders the development of state-of-the-art deep neural models. 

In response, in this paper we investigate how to make use of limited resources to improve MRC, using seven representative multiple-choice MRC datasets as case studies. As shown in Table~\ref{tab:related:answer}, the majority of the correct answer options in most of the datasets (except for ARC and MCTest) are non-extractive. Except for MultiRC, there is exactly one correct answer option for each question. For ARC and OpenBookQA, a reference corpus is provided instead of a single reference document associated with each question.

Here we give a formal \textbf{task definition}. Given a reference document $d$, a question $q$, and associated answer options $\{o_1, o_2, \dots, o_m\}$, the goal is to select the correct answer option(s). We can easily adapt our method to an MRC task that only provides a reference corpus (Section~\ref{sec:eval:five}).

%% file: 3_method.tex
\section{Approach}
\label{sec:method}

\begin{figure*}[htbp!]
   \begin{center}
   \includegraphics[width=0.98\textwidth]{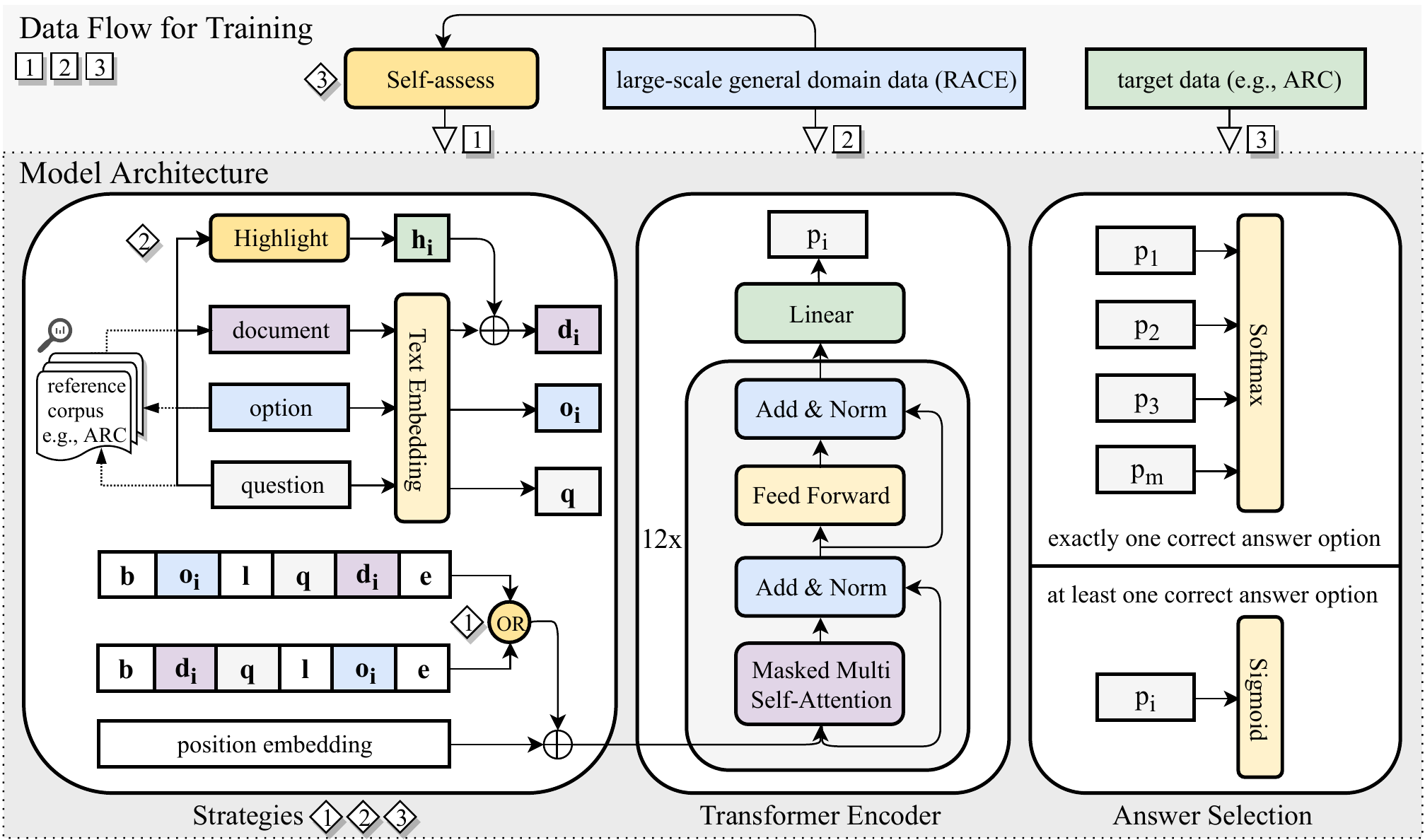}
   \end{center}
 \caption{Framework Overview. Strategy 1, 2, and 3 refer to back and forth reading (BF) (Section~\ref{sec:fb}), highlighting (HL) (Section~\ref{sec:hl}), and self-assessment (SA) (Section~\ref{sec:sa}), respectively.}
 \label{fig:overview}
\end{figure*}

We first introduce a neural reader based on a pre-trained transformer (Section~\ref{sec:trans}) and then elaborate on the strategies that are applied during the fine-tuning stage ---  back and forth reading (Section~\ref{sec:fb}), highlighting (Section~\ref{sec:hl}), and self-assessment (Section~\ref{sec:sa}).

\input{3.1_transformer.tex}
\input{3.2_strategies.tex}

%% file: 3.1_transformer.tex
\subsection{Framework Overview}
\label{sec:trans}

Our neural reader follows the framework of discriminatively fine-tuning a generative pre-trained transformer (GPT)~\cite{radfordimproving}. It adapts a pre-trained multi-layer transformer~\cite{vaswani2017attention,liu2018generating} language model to a labeled dataset $\mathcal{C}$, where each instance consists of a sequence of input tokens $x^1,\ldots,x^n$, along with a label $y$, by maximizing:
\begin{equation}
\small
\sum_{x,y} \log P (y\, | \,x^1,\ldots,x^n) + \lambda \cdot L(\mathcal{C})
\end{equation}
where $L$ is the likelihood from the language model, $\lambda$ is the weight of language model, and $P (y \, | \,  x^1,\ldots,x^n)$ is obtained by a linear classification layer over the final transformer block's activation of the language model. For multiple-choice MRC tasks, $x^1,\ldots,x^n$ come from the concatenation of a start token, a reference document, a question, a delimiter token, an answer option, and an end token; $y$ indicates the correctness of an answer option. We refer readers to~\newcite{radfordimproving} for more details.

Apart from placing a delimiter to separate the answer option from the document and question, the original framework pays little attention to task-specific structures in MRC tasks. Inspired by reading strategies, with limited resources and a pre-trained transformer, we propose three strategies to improve machine reading comprehension. We show the whole framework in Figure~\ref{fig:overview}. 

%% file: 3.2_strategies.tex
\subsection{Back and Forth Reading (BF)}
\label{sec:fb}

For simplicity, we represent the original input sequence of GPT during fine-tuning~\cite{radfordimproving} as [$dq \, \$ \, o$], where [, $\$$, and ] represent the start token, delimiter token, and end token, respectively. Inspired by back and forth reading, we consider both the original order and the reverse order [$o  \, \$\, qd$]. The token order within $d$, $q$, and $o$ is still preserved. We fine-tune two GPTs that use [$dq \, \$\, o$] and [$o  \, \$\, qd$] as the input sequence respectively, and then we ensemble the two models. We also consider other similar pairs of input sequences such as [$qd\, \$\,o$] and [$o\, \$\,dq$] in the experiments (Section~\ref{sec:eval:other_strategy}).

\subsection{Highlighting (HL)}
\label{sec:hl}
In the original implementation~\cite{radfordimproving}, during the fine-tuning stage of GPT, the text embedding of a document is independent of its associated questions and answer options. Inspired by highlights used in human reading, we aim to make the document encoding aware of the associated question-answer option pair ($q$, $o_i$). We focus on the content words in questions and answer options since they appear to provide more useful information~\cite{mirza2013ccg}, and we
identify them via their part of speech (POS) tags, one of: noun, verb, adjective, adverb, numeral, or foreign word.
Formally, we let $T$ be the set of POS tags of the content words. We let $\bm{d}$ denote the
sequence of the
text embedding of document $d$. We use $d^j$ to represent the $j^{th}$ token in $d$ and $\bm{d^j}$ to denote the text embedding of $d^j$. Given $d$ and a ($q$, $o_i$) pair, we define a \emph{highlight embedding} $\bm{h_i^j}$ for the $j^{th}$ token in $d$ as: 
\begin{equation}
\small
\bm{h_i^j} =
\begin{cases}
 \bm{\ell^{+}} &\text{if the POS tag of $d^j$ belongs to $T$,}   \\
 & \text{and $d^j$ appears in either $q$ or $o_i$}\\
 \bm{\ell^{-}} &\text{otherwise}
\end{cases} 
\label{eq:hi}
\end{equation}
\noindent where $\bm{\ell^{+}}$ and $\bm{\ell^{-}}$ are two trainable vectors of the same dimension as $\bm{d^j}$.

Following the above definition, the sequence of the highlight embedding $\bm{h_i}=\bm{h_i^1}, \bm{h_i^2}, \ldots, \bm{h_i^n}$ is of the same length as $\bm{d}$. We replace $\bm{d}$ with $\bm{d_i} = \bm{d} + \bm{h_i}$ when we encode a document. More specifically, we use the concatenation of $\bm{b}$, $\bm{d_i}$, $\bm{q}$, $\bm{l}$, $\bm{o_i}$, and $\bm{e}$ as the new input of GPT during fine-tuning (Section~\ref{sec:trans}), where $\bm{b}$, $\bm{l}$, and $\bm{e}$ denote the embedding of the start token, delimiter token, and end token, respectively, and $\bm{q}$ and $\bm{o_i}$ represent the sequence of the text embedding of $q$ and $o_i$, respectively.

\subsection{Self-Assessment (SA)}
\label{sec:sa}
While in previous work~\cite{radfordimproving}, the original GPT is directly fine-tuned on
an MRC end task, we instead develop a fine-tuning approach inspired by the self-assessment reading strategy.
In particular, we propose a simple method to generate questions and their associated multiple span-based answer options, which cover the content of multiple sentences from a reference document. By first fine-tuning a pre-trained model on these \emph{practice} instances, we aim to render the resulting fine-tuned model more 
aware of the input structure and to integrate information across multiple sentences as may be required to answer a given question. 

Concretely, we randomly generate no more than $n_{q}$ questions and associated answer options based on each document from the end task (\ie, RACE in this paper). We describe the steps as follows.

\vspace{-\topsep}
\begin{itemize}
\setlength\itemsep{-0.25em}
    \item \textbf{Input:}  a reference document from the end task.
    \item \textbf{Output:} a question and four answer options associated with the reference document.
    \item[1.] Randomly pick no more than $n_{s}$ sentences from the document and concatenate these sentences together.
    \item[2.] Randomly pick no more than $n_{c}$ non-overlapping spans from the concatenated sentences. Each span randomly contains no more than $n_{t}$ tokens within a single sentence. We concatenate the selected spans to form the correct answer option. We remove the selected spans from the concatenated sentences and use the remaining text as the question.
    \item[3.] Generate three distractors (\ie, wrong answer options) by randomly replacing spans in the correct answer option with randomly picked spans from the document.
\end{itemize}
\vspace{-\topsep}
where $n_q$, $n_s$, $n_c$, and $n_t$ are used to control the number and difficulty level of the questions.

%% file: 4_evaluation.tex
\section{Experiment}
\label{sec:experiment}

\subsection{Experiment Settings}
For most of the hyperparameters, we follow the work of~\newcite{radfordimproving}. We use the same preprocessing procedure and the released pre-trained transformer. We generate $119$k instances based on the reference documents from the training and development set of RACE~\cite{lai2017race}, with $n_q=10$, $n_s=3$, $n_c=4$, and $n_t=4$ (Section~\ref{sec:sa}). We first fine-tune the original pre-trained model on these automatically generated instances with $1$ training epoch (data flow $1$ boxed in Figure~\ref{fig:overview}). We then fine-tune the model on a large-scale general domain MRC dataset RACE with $5$ training epochs (data flow $2$ boxed in Figure~\ref{fig:overview}). Finally, we fine-tune the resulting model on one of the aforementioned six out-of-domain MRC datasets (at max $10$ epochs). See data flow $3$ boxed in Figure~\ref{fig:overview}. When we fine-tune the model on different datasets, we set the batch size to $8$, language model weight $\lambda$ to $2$. We ensemble models by averaging logits after the linear layer. For strategy highlighting (Section~\ref{sec:hl}), the content-word POS tagset $T=$ \{NN, NNP, NNPS, NNS, VB, VBD, VBG, VBN, VBP, VBZ, JJ, JJR, JJS, RB, RBR, RBS, CD, FW\}, and we randomly initialize $\bm{\ell^{+}}$ and $\bm{\ell^{-}}$.

\begin{table}[h!]
\centering
\scriptsize
\begin{tabular}{llcc}
\toprule
\multicolumn{2}{l}{\textbf{Approach}}   & \textbf{\#} & \textbf{RACE-M$|$RACE-H$|$RACE} \\ 
\midrule
\multicolumn{2}{l}{MMN~\cite{tang2019multi}}         & 9     & 64.7$~|~$55.5$~|~$58.2     \\
\multicolumn{2}{l}{GPT~\cite{radfordimproving}}      & 1     & 62.9$~|~$57.4$~|~$59.0     \\ 
\multicolumn{2}{l}{Human performance~\cite{lai2017race}} & 1     & 85.1$~|~$69.4$~|~$73.3     \\

\midrule
\multicolumn{2}{l}{\multirow{3}{*}{\shortstack[l]{$\text{GPT}^\star$}}}                                        
                                                     & 1    & 60.9$~|~$57.8$~|~$58.7                           \\
\multicolumn{2}{l}{}                                 & 2    & 62.6$~|~$58.4$~|~$59.6                           \\
\multicolumn{2}{l}{}                                 & 9    & 63.5$~|~$59.3$~|~$60.6                           \\
\midrule
\multirow{6}{*}{\begin{tabular}[l]{@{}l@{}}$\text{GPT}^\star$ \\ $~~~~$+ \\Strategies\end{tabular}} 
    & SA           & 1      & 63.2$~|~$59.2$~|~$60.4   \\
    & HL           & 1      & 67.4$~|~$61.5$~|~$63.2                           \\
    & BF           & 2      & 67.3$~|~$60.7$~|~$62.6                           \\
    & SA + HL      & 1      & \textbf{69.2}$~|~$\textbf{61.5}$~|~$\textbf{63.8} \\
    & SA + HL + BF & 2      & \textbf{70.9}$~|~$\textbf{63.2}$~|~$\textbf{65.4} \\
    & SA + HL + BF & 9      & \textbf{72.0}$~|~$\textbf{64.5}$~|~$\textbf{66.7} \\ 
\bottomrule
\end{tabular}
\caption{Accuracy ($\%$) on the test set of RACE (\#: number of (ensemble) models; SA: Self-Assessment; HL: Highlighting; BF: Back and Forth Reading; $^\star$: our implementation).} 
\label{tab:eval:race}
\end{table}

\subsection{Evaluation on RACE}
\label{sec:eval:race}
In Table~\ref{tab:eval:race}, we first report the accuracy of the state-of-the-art models (MMN and original fine-tuned GPT) and Amazon Turkers (Human performance). We then report the performance of our implemented fine-tuned GPT baselines and our models (GPT+Strategies). Results are shown on the RACE dataset~\cite{lai2017race} and its two subtasks: RACE-M collected from middle school exams and RACE-H collected from high school exams.

Our single and ensemble models outperform previous state-of-the-art (\ie, GPT and GPT ($9\times $)) by a large margin ($63.8\%$ \vs~$59.0\%$; $66.7\%$ \vs~$60.6\%$). 
The two single-model strategies -- self-assessment and highlighting -- improve over the single-model fine-tuned GPT baseline ($58.7\%$) by $1.7\%$ and $4.5\%$, respectively. 
Using the back and forth reading strategy, which involves two models, gives a $3.0\%$ improvement in accuracy compared to the ensemble of two original fine-tuned GPTs  ($59.6\%$). Strategy combination further boosts the performance. By combining self-assessment and highlighting, our single
model achieves a $5.1\%$ improvement in accuracy over the fine-tuned GPT baseline ($63.8\%$ \vs~$58.7\%$). We apply all the strategies by ensembling two such single models that read an input sequence in either the original or the reverse order, leading to a $5.8\%$ improvement in accuracy over the ensemble of two original fine-tuned GPTs ($65.4\%$ \vs~$59.6\%$). 

\begin{figure}[!h]
   \begin{center}
   \includegraphics[width=0.48\textwidth]{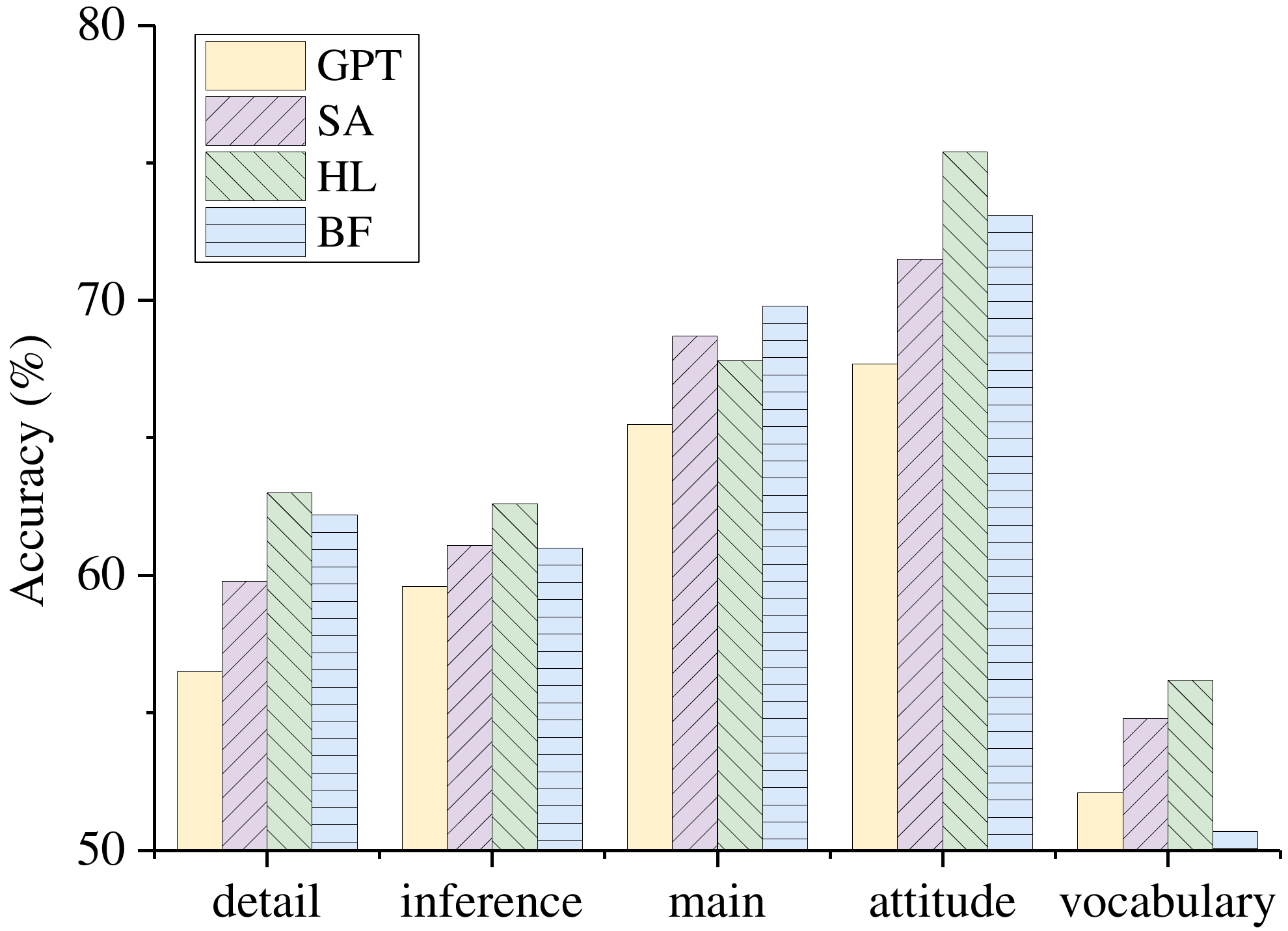}
   \end{center}
 \caption{Performance on different question types.}
 \label{fig:eval:question_type}
\end{figure}

To further analyze performance, we roughly divide the question types into five categories: detail (\emph{facts and details}), inference (\emph{reasoning ability}), main (\emph{main idea or purpose of a document}), attitude (\emph{author's attitude toward a topic or tone/source of a document}), and vocabulary (\emph{vocabulary questions})~\cite{qian2004evaluation,lai2017race} and annotate all the instances of the RACE development set. As shown in Figure~\ref{fig:eval:question_type}, compared to the fine-tuned GPT baseline, our single-model strategies (SA and HL) consistently improve the results across all categories. Compared to other strategies, highlighting is likely to lead to bigger gains for most question types.

Compared to human performance, there is still a considerable room for improvements, especially on RACE-M. We take a close look at the instances from the RACE-M development set that all our implementations fail to answer correctly. We notice that $82.0\%$ of them require one or multiple types of world knowledge (\eg, negation resolution, commonsense, paraphrase, and mathematical/logic knowledge~\cite{sugawara2017prerequisite,sugawara2017evaluation,sugawara2018makes}), especially when correct answer options are not explicitly mentioned in the reference document. For example, we need the knowledge --- \emph{the type of thing that is written by a writer can probably be a book} --- to answer the question \emph{``follow your heart is a \underline{~~~}''} from the context \emph{``Follow your heart by Andrew Matthews, an Australian writer, tells us that making our dreams real is life's biggest challenge''}. Besides, $19.7\%$ of these failed instances require coreference resolution. It might be promising to leverage coreference resolvers to connect nonadjacent relevant sentences.

\begin{table*}[h!]
\centering
\scriptsize
\begin{tabular}{p{2.1cm}llp{0.4cm}cccc}
\toprule
\bf Task  & \bf Metric   &  \multicolumn{2}{c}{\bf Previous STOA} & \bf GPT & \bf GPT (2$\times$) & \bf GPT+Strategies & \bf GPT+Strategies (2$\times$) \\
\midrule
ARC-Easy     & Acc. & \newcite{clark2018think}  & 62.6 & 57.0    &  57.1  & 66.6  & 68.9       \\
ARC-Challenge  &  Acc.  & \newcite{ni2018learning}  & 36.6  & 38.2  & 38.4  & 40.7  & 42.3     \\
OpenBookQA  &  Acc.  & \newcite{mihaylov2018can} & 50.2  & 52.0  & 52.8  & 55.2  & 55.8        \\
MCTest-MC160  &  Acc.  & \newcite{chung2017supervised}  & 76.4 & 65.4  & 65.8  & 80.0  & 81.7    \\
MCTest-MC500  &  Acc.  & \newcite{chung2017supervised}  & 72.3 & 61.5  & 61.0  & 78.7  & 82.0    \\
SemEval     &  Acc.  & \newcite{chen2018hfl}          & 84.1 & 88.0  & 88.6  & 88.8  & 89.5    \\
ROCStories  &  Acc.  & \newcite{radfordimproving}     & 86.5  & 87.1  & 87.5  & 88.0  & 88.3    \\
\multirow{3}{*}{MultiRC} &$\text{F1}_{m}$ & \newcite{khashabi2018looking} 
                                                      & 66.5  & 69.3  & 70.3  & 71.5  & 73.1   \\
                         & $\text{F1}_a$  & \newcite{khashabi2018looking}             
                                                      & 63.2  & 67.2  & 67.7  & 69.2  & 70.5   \\
                         &  Acc.$^\dagger$  & \newcite{khashabi2018looking}              
                                                      & 11.8  & 15.2  & 16.5  & 22.6  & 21.8    \\
\midrule
\bf Average              & Acc.     &                 & 60.1  & 58.1  & 58.5  & \bf 65.1  & \bf 66.3    \\             
\bottomrule
\end{tabular}
\caption{Performance (\%) on the test sets of ARC, OpenBookQA, MCTest, SemEval-2018 Task 11, and ROCStories and the development set of MultiRC (Acc.: Accuracy; $\text{F1}_m$: macro-average F1; $\text{F1}_a$: micro-average F1; $^\dagger$: using the joint exact match accuracy (\ie, $\text{EM}_0$ reported by the official evaluation~\cite{khashabi2018looking})). RACE is used as the source task for all our implementations.} %
\label{tab:eval:transfer}
\end{table*}

\begin{table*}[h!]
\scriptsize
\centering
\begin{tabular}{lcccccc|c}
\toprule
\multirow{2}{*}{\bf Approach}     & \bf ARC   & \bf OpenBookQA  & \bf MCTest & \bf SemEval  & \bf ROCStories & \bf MultiRC & \bf Average \\

                     & Easy $|$ Challenge    & - & MC160 $|$ MC500   & -            & -  & - & - \\
               & Acc.      & Acc.     & Acc.   & Acc.  & Acc. & $\text{F1}_{m}$ $|$ $\text{F1}_{a}$ $|$ Acc.$^\dagger$ & Acc. \\
\midrule
GPT   & 54.0 $|$ 30.3   & 50.0 & 58.8 $|$ 52.0   & 87.3 & 86.7 & 69.3 $|$ 66.2 $|$ 11.9 & 53.9 \\
GPT (2$\times$)  & 53.9 $|$ 30.7 &  50.0 & 60.0 $|$ 54.0 & 88.0  & 87.0 &  69.3 $|$ 66.5 $|$ 13.1 & 54.6 \\
GPT+Strategies  &  61.9 $|$ 35.0 &  54.2 & 67.5 $|$ 64.7 &87.6  &87.4 &68.8	$|$ 67.4  $|$ 16.2 & \bf 59.3 \\
GPT+Strategies (2$\times$)  &  63.1 $|$  35.4  &   55.0 &  70.8 $|$  64.8  &  88.1 &  88.1 &   69.7 $|$  67.9 $|$  16.9 & \bf 60.3 \\
\bottomrule
\end{tabular}
\caption{Performance (\%) on the test sets of ARC, OpenBookQA, MCTest, SemEval-2018 Task 11, and ROCStories and the development set of MultiRC using the target data only (\ie, without the data flow 1 and 2 boxed in Figure~\ref{fig:overview}) (Acc.: Accuracy; $\text{F1}_m$: macro-average F1; $\text{F1}_a$: micro-average F1; $^\dagger$: using the joint exact match accuracy (\ie, $\text{EM}_{0}$ reported by the official evaluation~\cite{khashabi2018looking})). %
\label{tab:eval:notransfer}}
\end{table*}

\subsection{Further Discussions on Strategies}
\label{sec:eval:other_strategy}

Besides the strategies introduced in Section~\ref{sec:method}, we also explore other reading strategies such as {\sc summarization} (\emph{``I take an overall view of the text to see what it is about before carefully reading it.''}) by appending an extractive summary~\cite{boudin2015concept} before each reference document, which is shown less effective for machine reading comprehension in our experiments compared to the strategies we focus on. In this section, we further discuss the three strategies.

\textbf{Back and Forth Reading} We notice that the input order difference between two ensemble models is likely to yield performance gains. Besides ensembling two models that use input sequence [$dq \, \$\, o$] and [$ o\, \$\,qd$] respectively, we also investigate other reverse or almost reverse pairs. For example, we can achieve better results by ensembling [$qd \, \$\, o$] and [$o\, \$\,dq$] ($61.0\%$) or [$qd \, \$\, o$] and [$o\, \$\,qd$] ($61.7\%$), compared to the ensemble of two original fine-tuned GPTs (both of them use [$d\, \$\,qo$]) on the RACE dataset ($59.6\%$ in Table~\ref{tab:eval:race}).

\textbf{Highlighting} We try two variants to define highlight embeddings (Equation~\ref{eq:hi} in Section~\ref{sec:hl}) by considering the content of questions only or answer options only. Experiments show that using partial information yields a decrease in accuracy ($60.6\%$ and $61.0\%$, respectively) compared to $63.2\%$ (Table~\ref{tab:eval:race}), achieved by considering the content words in a question and its answer options. We attempt to also highlight the coreferential mentions of the content words, which does not lead to further gains, though.

\textbf{Self-Assessment} We explore alternative approaches to generate questions. For example, we use the Wikipedia articles from SQuAD~\cite{rajpurkar2016squad} instead of the general domain documents from the end task RACE. We generate the same number of questions as the number of questions we generate using RACE following the same steps mentioned in Section~\ref{sec:sa}. Experiments show that this method also improves the accuracy of the fine-tuned GPT baseline ($59.7\%$ \vs~ $58.7\%$).
As self-assessment can be somehow regarded as a data augmentation method, we investigate other unsupervised question generation methods such as sentence shuffling and paraphrasing via back-translation~\cite{ding2018ynu,yu2018qanet}. Our experiments demonstrate that neither of them results in performance improvements on the RACE dataset.

\subsection{Adaptation to Other Non-Extractive Machine Reading Comprehension Tasks}
\label{sec:eval:five}
We follow the philosophy of transferring the knowledge from a high-performing model pre-trained on a large-scale supervised data of a source task to a target task, in which only a small amount of training data is available~\cite{chung2017supervised}. RACE has been used to pre-train a model for other MRC tasks as it contains the largest number of general domain non-extractive questions (Table~\ref{tab:related:answer})~\cite{ostermann2018semeval,wang2018yuanfudao}. In our experiment, we also treat RACE as the source task and regard six representative non-extractive multiple-choice MRC datasets from multiple domains as the target tasks.

We require some task-specific modifications considering the different structures of these datasets. In ARC and OpenBookQA, there is no reference document associated with each question. Instead, a reference corpus is provided, which consists of unordered science-related sentences relevant to questions. We therefore first use Lucene~\cite{lucene} to retrieve the top $50$ sentences by using the non-stop words in a question and one of its answer options as a query. The retrieved sentences are used to form the reference document for each answer option. In MultiRC, a question could have more than one correct answer option. Therefore, we use a sigmoid function instead of softmax at the final layer (Figure~\ref{fig:overview}) and regard the task as a binary (\ie, correct or incorrect) classification problem over each (document, question, answer option) instance. When we adapt our method to the non-conventional MRC dataset ROCStories, which aims at choosing the correct ending to a four-sentence incomplete story from two answer options~\cite{mostafazadeh2016corpus}, we leave the question context empty as no explicit questions are provided. Since the test set of MultiRC is not publicly available, we report the performance of the model that achieves the highest micro-average F1 ($\text{F1}_a$) on the development set. For other tasks, we select the model that achieves the highest accuracy on the development set and report the accuracy on the test set. %

We first fine-tune GPT using our proposed three strategies on RACE and further fine-tune the resulting model on one of the six target tasks (see Table~\ref{tab:eval:transfer}). During the latter fine-tuning stage, besides the \emph{highlighting} embeddings inherited from the previous fine-tuning stage, we also apply the strategy \emph{back and forth reading}, and we do not consider \emph{self-assessment} since the model has already benefited from the high-quality RACE instances during the first fine-tuning stage. We compare with the baselines that are first fine-tuned on RACE and then fine-tuned on a target task without the use of strategies, which already outperform previous state-of-the-art (SOTA) on four out of six datasets (OpenBookQA, SemEval-2018 Task 11, ROCStories, and MultiRC). By using the strategies, we obtain a $7.8\%$ absolute improvement in average accuracy over the ensemble baseline ($58.5\%$) and a $6.2\%$ absolute improvement over previous SOTA ($60.1\%$). 

To further investigate the contribution of the strategies, we directly fine-tune GPT on a target task without using the labeled data in RACE (\ie, we only keep data flow $3$ in Figure~\ref{fig:overview}). Compared to the baseline that is fine-tuned without using strategies ($54.6\%$), we obtain a $10.4\%$ relative improvement in average accuracy ($60.3\%$) and especially large improvements on datasets ARC, OpenBookQA, and MCTest (Table~\ref{tab:eval:notransfer}).

%% file: 5_relatedwork.tex
\section{Related Work}
\label{sec:related}

\subsection{Methods for Multiple-Choice Machine Reading Comprehension}

We primarily discuss methods applied to large-scale datasets such as RACE~\cite{lai2017race}. Researchers develop a variety of methods with attention mechanisms~\cite{chen2016thorough,dhingra2017gated,dfn2018,tay2018multi,tang2019multi} for improvement such as adding an elimination module~\cite{parikh2018eliminet} or applying hierarchical attention strategies~\cite{haf2017,wang2018co}. These methods seldom take the rich external knowledge (other than pre-trained word embeddings) into considerations. Instead, we investigate different strategies based on an existing pre-trained transformer~\cite{radfordimproving} (Section~\ref{sec:trans}), which leverages rich linguistic knowledge from external corpora and achieves state-of-the-art performance on a wide range of natural language processing tasks including machine reading comprehension.  %

\subsection{Transfer Learning for Machine Reading Comprehension and Question Answering}
Transfer learning techniques have been successfully applied to machine reading comprehension~\cite{golub2017two,chung2017supervised} and question answering~\cite{min2017question,wiese2017neural}. Compared to previous work, we simply fine-tune our model on the source data and then further fine-tune the entire model on the target data. The investigation of methods such as adding additional parameters or an L2 loss and fine-tuning only part of the parameters is beyond the scope of this work.

\subsection{Data Augmentation for Machine Reading Comprehension Without Using External Datasets}
Previous methods augment the training data for extractive machine reading comprehension and question answering by randomly reordering words or shuffling sentences~\cite{ding2018ynu,li2018lyb3b} or generating questions through paraphrasing~\cite{yang2017semi,yuan2017machine}, which require a large amount of training data or limited by the number of training instances~\cite{yu2018qanet}. In comparison, our problem (\ie, question and answer options) generation method does not rely on any existing questions in the training set, and the generated questions can involve the content of multiple sentences in a reference document.

%% file: 6_conclusion.tex
\section{Conclusions}

Inspired by previous research on reading strategies for improved comprehension levels of human readers, we propose three strategies (\ie, back and forth reading, highlighting, and self-assessment), aiming at improving machine reading comprehension using limited resources: a pre-trained language model and a limited number of training instances. By applying the proposed three strategies, we obtain a $5.8\%$ absolute improvement in accuracy over the state-of-the-art performance on the RACE dataset. By fine-tuning the resulting model on a new target task, we achieve new state-of-the-art results on six representative non-extractive MRC datasets from multiple domains that require a diverse range of reading skills. These results consistently indicate the effectiveness of our proposed strategies and the general applicability of our fine-tuned model that incorporates these strategies. 